\documentclass{article}

\usepackage{PRIMEarxiv}

\usepackage[utf8]{inputenc} 
\usepackage[T1]{fontenc}    
\usepackage{hyperref}       
\usepackage{url}            
\usepackage{booktabs}       
\usepackage{amsfonts}       
\usepackage{nicefrac}       
\usepackage{microtype}      
\usepackage{lipsum}
\usepackage{fancyhdr}       
\usepackage{graphicx}       
\usepackage{amsmath}
\graphicspath{{media/}}     

\usepackage{algorithm}%
\usepackage[noend]{algpseudocode}%

\title{An Accuracy-Aware Extension to LRP-Based Pruning for CNNs
to Prevent Cascading Accuracy Degradation in Data-Scarce Transfer Learning
}

\author{
  Daisuke Yasui*, Toshitaka Matsuki, Hiroshi Sato \\
  Mathematics and Computer Science \\
  National Defense Academy of Japan \\
  Yokosuka, Japan\\
  \texttt{*ed24003@nda.ac.jp} \\
}

\begin{document}
\fancyfoot[L]{\small Published in \textit{Scientific Reports}, 2026. 
DOI: \href{https://doi.org/10.1038/s41598-026-47992-8}
{10.1038/s41598-026-47992-8}}
\maketitle

\begin{abstract}
Convolutional Neural Networks (CNNs) pre-trained on large-scale datasets such as ImageNet are widely used as feature extractors to construct high-accuracy classification models from scarce data for specific tasks.
In such scenarios, fine-tuning the pre-trained CNN is difficult due to data scarcity, necessitating the use of fixed weights.
However, when the weights are kept fixed, many filters that do not contribute to the target task remain in the model, leading to unnecessary redundancy and reduced efficiency.
Therefore, effective methods are needed to reduce model size by pruning filters that are unnecessary for inference.
To address this, approaches utilizing Layer-wise Relevance Propagation (LRP) have been proposed. 
LRP quantifies the contribution of each filter to the inference result, enabling the pruning of filters with low relevance.
However, existing LRP-based pruning methods have been observed to cause cascading accuracy degradation.
In this study, we introduce an accuracy-aware pruning control mechanism for existing LRP-based filter pruning methods, which suppresses cascading accuracy degradation by dynamically adjusting the pruning rate and the pruning order using the harmonic mean of class accuracy, and compresses the pre-trained model while preserving task-specific performance in a small-data environment.
We demonstrate that this control mechanism effectively mitigates cascading accuracy degradation and achieves higher classification accuracy compared to existing LRP-based pruning methods, improving the class-averaged area under the accuracy-pruning-rate curve (AUC) of VGG16 by approximately 15\% over conventional LRP-based approaches.
\end{abstract}


\section{Introduction}
Convolutional neural networks (CNNs) have demonstrated outstanding performance in image classification tasks and are utilized in diverse applications\cite{CNN_classification, CNN_anomalydetection}.
In domains where training data is scarce due to acquisition difficulties, transfer learning is widely adopted to use knowledge acquired from a source task with a large-scale dataset for a target task \cite{TL_FL1, TL_FL2}.
A common approach employs an ImageNet-pretrained CNN as a feature extractor, classifying the extracted features via fully connected layers. \cite{FL_TL, FL_TL2}.

However, the whole representational capabilities acquired in the source task are not necessarily essential for the target task and often include redundancies \cite{redundant}.
This redundancy increases computational costs.
In particular, convolutional layers account for the majority of the inference cost in CNNs  \cite{CNN_calc_cost}.
For example, in VGG16\cite{vgg}, the majority of operations are concentrated in convolutional layers, requiring billions of FLOPs to classify a single image \cite{Lan2021Compressing}.
Such high computational costs pose a significant obstacle to implementation in resource-constrained environments such as embedded systems, autonomous agents, and mobile devices \cite{Chen2020DeepMobileEmbedded}.
Consequently, reducing model size by pruning redundant filters while maintaining generalization performance for the target task is essential \cite{YEOM}.

Traditional CNN pruning methods that prune filters regardless of the target task (e.g., pruning filters with small absolute weights \cite{pruning_with_weight}) typically require retraining to compensate for accuracy loss.
However, in data-scarce environments, retraining carries the risk of causing overfitting, which conversely can detrimentally affect generalization performance \cite{Molchanov2017Pruning}.
Therefore, pruning in data-scarce environments requires a method capable of accurately identifying and removing redundant filters non-contributory to the target task, without relying on retraining.
Layer-wise Relevance Propagation (LRP) is a method that quantifies the influence of individual filters on inference results by propagating relevance from the output layer to the input layer \cite{LRP}.
Consequently, LRP-based criteria are considered effective for filter pruning in data-scarce environments \cite{YEOM, PXP}.

However, existing LRP-based pruning methods suffer from a significant limitation.
When the classification accuracy of a particular class deteriorates by pruning, the relevance scores of the filters contributing to that class decrease correspondingly.
This causes a problem where filters essential for the inference of the class become targets for deletion sequentially due to diminished relevance score, leading to cascading accuracy degradation.
As a result, functionalities essential for the target task are severely compromised.

To address this issue, we introduce DPX-SD (Dynamic Pruning by eXplain for Scarce Data), which extends existing LRP-based pruning methods by incorporating an accuracy-aware control mechanism.
DPX-SD regulates the pruning process by dynamically adjusting the pruning rate and re-evaluating the filter deletion order based on class-wise accuracy balance in addition to conventional relevance scores. 
Specifically, when the harmonic mean of class accuracies decreases after a pruning step, the additional pruning rate is reduced, and if the balance cannot be restored, the pruning order is re-evaluated to avoid removing filters that preserve minority-class performance.
Through this accuracy-guided control, cascading accuracy degradation is suppressed while model compression is achieved without modifying the underlying relevance computation, enabling stable task performance even in data-scarce environments.

To verify the effectiveness of the proposed accuracy-aware pruning control framework, we conducted comparison experiments with existing LRP and weight based pruning methods \cite{YEOM, PXP, pruning_with_weight}.
Following the settings of existing research, we used only a few dozen images for randomly selected subclasses from ImageNet and performed pruning while maintaining the accuracy of the sub-tasks in each method.
This setting simulates scenarios where a model constructed via transfer learning is compressed post-hoc in domains with data acquisition difficulties, such as medical imaging \cite{Schouten2021Tens} or satellite imagery \cite{Bi2023Contrastive}.

The main contributions of this study are summarized as follows:
\begin{itemize}
    \item We analytically identified the mechanism of cascading accuracy degradation in existing LRP-based pruning approaches, where class-wise accuracy imbalance distorts relevance rankings and leads to the sequential removal of task-critical filters.
    \item We introduce DPX-SD as an accuracy-aware pruning control framework that regulates the pruning process without modifying the underlying relevance computation. 
    Specifically, the framework integrates two complementary control mechanisms: 
    (i) a \textit{pruning-rate adjustment mechanism} that dynamically restricts additional pruning based on the harmonic mean of class accuracies, and 
    (ii) a \textit{pruning-order re-evaluation mechanism} that temporarily protects filters preserving minority-class performance when accuracy balance deteriorates.
    \item Through extensive evaluations under data-scarce transfer-learning conditions using ImageNet subclasses, we demonstrate that the proposed control framework consistently improves both accuracy retention and compression efficiency compared with conventional LRP-based pruning strategies.
\end{itemize}

The structure of this paper is as follows.
Section 2 outlines related work on transfer learning, model compression,
and LRP-based pruning methods.
Section 3 details the problems with existing methods and describes the proposed method as a solution to them.
Section 4 evaluates the accuracy and compression performance of the proposed method through comparative experiments.
Finally, Section 5 presents the conclusion of this study and future prospects.
\section{Related Work}
This section reviews related work on transfer learning, model compression, and relevance-based pruning methods.
First, we review prior studies on transfer learning and training-based model compression approaches.
Next, we summarize relevance-based pruning methods that estimate filter importance without weight updates when retraining is impractical due to the risk of overfitting, focusing on Layer-wise Relevance Propagation (LRP).

\subsection{Transfer Learning and Model Compression}
Convolutional neural networks have demonstrated strong performance across a wide range of application domains by transferring weights pre-trained on large-scale datasets to new target tasks.
Recent studies have also demonstrated the effectiveness of transfer learning in practical remote-sensing applications; for example, accuracy improvements have been reported in LEO satellite-oriented wildfire detection by leveraging pre-trained deep neural networks\cite{satellite}.
In domains where annotation costs are high, such as medical image analysis and satellite image interpretation, it is common practice to utilize pre-trained models as feature extractors.
However, these pre-trained networks often contain a large number of parameters and filters, which leads to high computational cost and memory consumption during inference.
This inefficiency arises because the entire network is utilized even though many filters are not essential for the target task.
To address this issue, model compression and pruning techniques have been studied.

Most conventional compression methods integrate pruning into the training process and rely on retraining to preserve performance.
More recently, advanced approaches have been proposed that explicitly control the learning dynamics during compression.
Representative examples include curriculum-based compression that gradually increases pruning difficulty and reinforcement-guided knowledge transfer frameworks that optimize pruning policies to mitigate accuracy loss \cite{curricurum}.
Nevertheless, these approaches generally assume the availability of sufficient training data and repeated optimization cycles. 
In many real-world scenarios, however, only limited data are available, which increases the risk of overfitting during additional retraining. 
A related line of work addresses data imbalance by alternating between imbalanced and dynamically balanced datasets to stabilize learning and prevent minority-class collapse \cite{imbalance}. 
However, such strategies still rely on iterative retraining and may be insufficient when the overall amount of training data is small. 
In such situations, it becomes crucial to compress models without retraining.

To enable effective post-hoc compression, it is necessary to quantitatively evaluate how much each filter or feature contributes to the target task.
Among various techniques for estimating such importance without modifying model weights, Layer-wise Relevance Propagation (LRP) has been widely adopted as a representative approach, and it serves as the foundational framework in this study.

\subsection{Relevance-based Pruning Methods}
    In this subsection, we first describe Layer-wise Relevance Propagation (LRP) as a method for estimating filter importance, and then review existing pruning approaches that utilize LRP.
    
    \textbf{LRP.}
        Layer-wise Relevance Propagation (LRP) quantifies the contribution of each input feature or intermediate layer unit to the neural network's output in terms of "relevance". 
        First, in forward propagation, the activation value $a_j^{l}$ of neuron $j$ in layer $l$ is calculated as follows:
        \begin{equation}
            a_j^{l} = \sigma\left(\sum_i a_i^{l-1} w_{i,j} + b_j\right),
        \end{equation}
        where $i$ and $j$ denote the node indices within their respective layers, $w_{i,j}$ is the weight, $b_j$ is the bias, and $\sigma(\cdot)$ represents the nonlinear activation function.
        Subsequently, in LRP, a scalar relevance value $R$ is assigned to the target output class and back-propagated through the layers.
        While various propagation rules exist, the $\epsilon$-rule is widely recognized as effective for relevance-based pruning \cite{PXP}.
        In the $\epsilon$-rule, the relevance $R_{i\leftarrow j}^{l-1}$ propagated from node $j$ in layer $l$ to node $i$ in layer $l-1$ is given by the following equation:
        \begin{equation}\label{eq:LRP_rule}
            R_{i\leftarrow j}^{l-1} =
            \frac{a_i^{l-1} w_{i,j}}
                  {\sum_{i'} a_{i'}^{l-1} w_{i',j} + \epsilon} 
            \, R_j^{l} ,
        \end{equation}
        Here, the second term in the denominator serves as a stabilizer.
        The parameter $\epsilon$ is a small value introduced to prevent division by zero or excessive scaling.
        $i'$ represents the indices of all nodes belonging to the same layer $l-1$.
        This equation dictates that the relevance $R_j^{l}$ from the upper layer is distributed proportionally to the contribution of the signal $a_i^{l-1} w_{i,j}$ from each node in layer $l-1$.
        Thus, the $\epsilon$-rule assigns relevance by accounting for the forward propagation structure, thereby quantifying the contribution of each node or filter to the network output.
    
\begin{figure}[ht]
            \centering
            \includegraphics[width=0.9\columnwidth, page=1, trim={40mm 15mm 30mm 0mm}, clip]{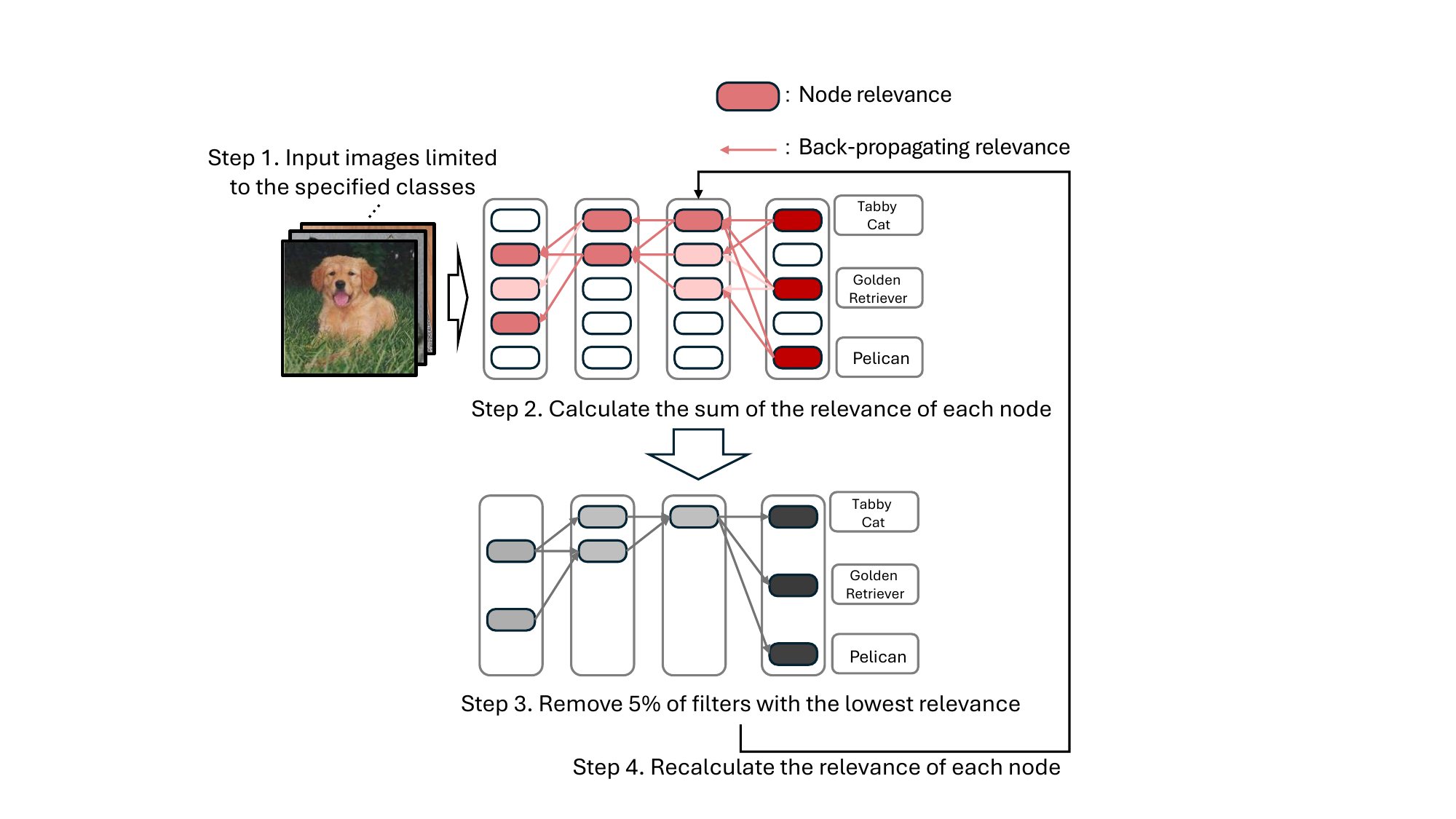}
            \caption[Calculation flow of pruning by LRP]
            {Calculation flow of pruning by LRP}
            \label{Fig:base_line}
        \end{figure}

    \textbf{LRP-based Pruning Methods.}
        Recent studies have explored utilizing feature relevance scores to prune pre-trained models while preserving domain-specific classification capabilities.
        Representative works in this direction involve calculating filter relevance via LRP and sequentially removing filters with minimal contribution to the target class decision \cite{YEOM, PXP}.
        In LRP-based pruning, using a small amount of data for the specific domain classification task, the relevance of all filters is calculated using LRP, and filters with low relevance are removed at a fixed rate.
        Hatef et al. \cite{PXP} compute the relevance of each filter once and perform pruning in ascending order of relevance.
        However, removing filters inevitably alters relevance propagation.
        Yeom et al. \cite{YEOM} propose an iterative method where filter relevance is calculated, the bottom 5\% are removed, and relevance is subsequently recalculated.
        Based on the updated ranking, an additional 5\% of filters are removed, and this process repeats.
        Fig. \ref{Fig:base_line} shows the flowchart of the method by Yeom et al. \cite{YEOM}.
        First, the relevance of each filter is calculated and summed when a small number of reference images limited to a specific group of classes are input (Steps 1 to 2 in Fig. \ref{Fig:base_line}; the color intensity represents the magnitude of relevance).
        Then, a ranking of the total relevance is created, and the lowest-ranked filters are removed (Step 3 in Fig. \ref{Fig:base_line}).
        Subsequently, the relevance is calculated again in that model (Step 4 in Fig. \ref{Fig:base_line}).
        However, this method leaves room for improvement in maintaining domain-specific accuracy during pruning, as discussed in Sec. \ref{sec:Proposed_method_section}.
\section{Proposed Method}\label{sec:Proposed_method_section}
    In this work, we propose DPX-SD (Dynamic Pruning by eXplain for Scarce Data).
    The central idea is not to modify the relevance computation itself, but to introduce an accuracy-aware control mechanism that regulates both the pruning rate and the selection of filters.
    In conventional LRP-based pruning, filters are removed solely according to relevance ranking. 
    Although relevance effectively captures contribution to inference, relying on this ranking alone can unintentionally eliminate filters that are critical for minority classes, thereby triggering cascading accuracy degradation.
    This is discussed in Sec.\ref{sec:factor_countermeasure}.

    To address this issue, the proposed pruning control mechanism monitors class-wise accuracy after each pruning step and activates corrective mechanisms only when a degradation is detected.
    These mechanisms form a feedback loop consisting of two complementary controls: (i) adaptive reduction of the additional pruning rate, and (ii) temporary protection of candidate
    filters. 
    To clarify the overall behavior of this feedback-controlled pruning process, Algorithm~\ref{alg:dpxsd} presents a simplified high-level procedure.
    However, to better understand the necessity of such accuracy-aware control, Sec.~\ref{sec:factor_countermeasure} analyzes the underlying factors of cascading accuracy degradation, while Sec.~\ref{sec:Proposed_method} presents the corresponding countermeasures.

    \begin{algorithm}[h]
        \caption{High-level procedure of the proposed accuracy-aware pruning control. Detailed steps are provided in the Supplementary Information.}
        \label{alg:dpxsd}
        \begin{algorithmic}[1]
            \State \textbf{Input:} CNN Model $M$, reference images $X_r$, pruning step $P_{step}$\%
            \State Evaluate class-wise accuracy and harmonic mean $A$
            \While{pruning rate $< P_{max}$}
                \State Compute filter relevance $R$ by $X_r$
                \State Prune bottom $P_{step}\%$ filters by $R$
                \State Evaluate class-wise accuracy and harmonic mean $A'$ by $X_r$
                \If{$A' < A$}
                    \State $P_{step} \leftarrow P_{step} / 2$
                    \If{$P_{step}$ corresponds to one filter}
                        \State Re-evaluate candidates (up to $T_{max}$ trials)
                        \State Select the optimal filter to prune that maximizes $A'$
                    \EndIf
                \EndIf
                \State $A \leftarrow A'$
            \EndWhile
            \State \textbf{Output:} Pruned model $M'$
        \end{algorithmic}
    \end{algorithm}

    \subsection{Reasons for the Inability of Existing Methods to Sufficiently Prune While Maintaining Accuracy}\label{sec:factor_countermeasure}
        Deterioration in class-wise accuracy balance causes relative undervaluation of relevance scores for filters contributing to under-performing classes.
        Suppose that pruning causes the accuracy balance among classes to deteriorate, resulting in a relative decline in the accuracy of class $C$ compared to other classes.
        A decline in accuracy for class $C$ suggests insufficient signal propagation to the corresponding output node.
        Consequently, relevance originating from class $C$'s output node is diminished during backpropagation to the input layer.
        Thus, the relevance of filters contributing to the output of class~$C$ becomes relatively low, and as a result, the scaling of relevance with respect to accuracy can cease to be uniform across filters.
        This reduction in the relevance of filters contributing to low-accuracy classes, caused by the deterioration of inter-class accuracy balance, induces two major problems in pruning.
        
        The first issue is a cascading decline in accuracy for already compromised classes.
        Due to the deterioration of accuracy balance, filters contributing to low-accuracy classes are assigned lower relevance scores and become more likely to be removed in the next pruning step, thereby creating a cycle where the accuracy of these classes declines further.
        To suppress this phenomenon, it is necessary to proceed with pruning in a way that maintains a high average accuracy across all classes while preventing the deterioration of inter-class accuracy balance, which serves as the starting point of this chain reaction.
        We introduce the harmonic mean of class accuracies, denoted as $A$, to simultaneously assess inter-class balance and overall performance.
        Let $A_c$ be the accuracy of class $c$; then $A$ is defined by the following equation:
        \begin{equation}
            A = \begin{cases}
                \dfrac{|C|}{\sum_{c \in C} A_{c}^{-1}} & \text{if } A_c > 0 \ \forall c \in C, \\[6pt]
                0 & \text{otherwise.}
            \end{cases}
            \label{eq:halmo_mean_A}
        \end{equation}
        For example, in binary classification, if $(A_\alpha, A_\beta) = (0.90, 0.90)$, then $A \approx 0.90$. 
        However, in the case of $(1.00, 0.80)$, where the arithmetic mean is the same, $A$ drops to approximately $0.89$.
        Furthermore, if $(A_\alpha, A_\beta) = (1.00, 0.00)$, then $A = 0$ since $A_\beta^{-1}$ is undefined.
        Thus, $A$ effectively captures both average accuracy and class distribution balance.
        A higher $A$ after pruning indicates superior average accuracy and balance among classes, whereas a decrease in $A$ after pruning suggests a decline in overall accuracy or class balance.
        As the total pruning rate increases, the representational capacity of the model diminishes, making a decline in $A$ inevitable.
        At this stage, we first reduce the additional pruning rate to a point where $A$ can be restored to its value before the decline, thereby recovering the model's representational capacity, and then recalculate the relevance scores.
        This enables pruning by evaluating relevance within a network structure that maintains inter-class accuracy balance, thereby preventing the chain reaction of accuracy collapse in specific classes (Additional function 1).
        
        However, as the pruning rate increases further and the model's representational capacity declines more, there is a possibility that $A$ will decrease and the deterioration of inter-class accuracy balance will be unavoidable, even if the additional pruning rate is reduced to the minimum unit (removing only the single filter with the lowest relevance).
        At this point, the second problem arises: there can be filters whose relevance is not small, but whose removal has little effect on accuracy.
        When inter-class accuracy balance deteriorates, discrepancies arise across filters regarding the scale of relevance relative to accuracy preservation. 
        Consequently, among low-relevance filters, the ranking of scores fails to accurately reflect the actual contribution to accuracy.
        Pruning filters in order of relevance in this state risks overlooking filters that have a smaller impact on $A$.
        Therefore, it is necessary to temporarily prune each of the low-relevance candidate filters, measure the actual $A$, and exploratively identify the filter that minimizes the impact (Additional function 2).
        
        Based on these considerations, this study proposes adding the following two functions to existing LRP-based pruning methods:
        \begin{itemize}
            \item \textbf{Change of pruning rate:} A function that dynamically restricts the additional pruning rate to a range where the harmonic mean $A$ can be maintained.
            \item \textbf{Change of pruning order:} A function that searches for and deletes the filter that actually maximizes $A$ when $A$ cannot be maintained even if the additional pruning rate is reduced to the minimum unit (one filter).
        \end{itemize}
    
    \subsection{Proposed Method Based on Countermeasures (DPX-SD: Dynamic Pruning by eXplain for Scarce Data)}\label{sec:Proposed_method}
\begin{figure}[H]
            \centering
            \includegraphics[width=1\columnwidth, page=2, trim={25mm 15mm 60mm 0mm}, clip]{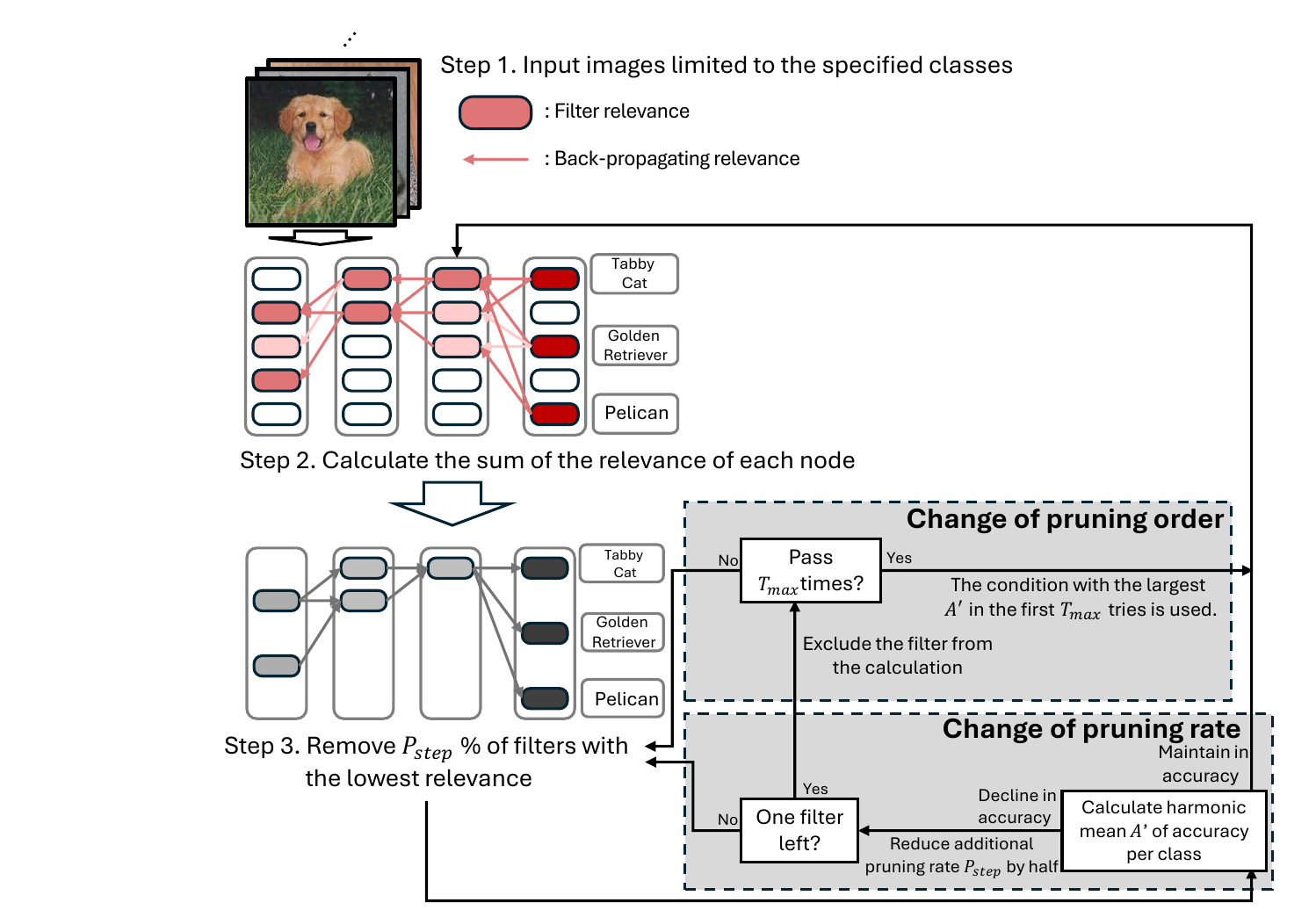}
            \caption
            {DPX-SD (Dynamic Pruning by eXplain for Scarce Data) is an extension of the LRP-based pruning method, augmented with two additional pruning control mechanisms that modify the pruning order and the pruning rate.}
            \label{Fig:improved_method}
        \end{figure}

        Building on the analysis in Sec. \ref{sec:factor_countermeasure}, Fig. \ref{Fig:improved_method} illustrates the proposed framework, improving upon the baseline in Fig. \ref{Fig:base_line} (represented as pseudo-code in Supplementary Information).
        Following the experimental conditions of Yeom et al.\cite{YEOM} and Hatef et al.\cite{PXP}, the proposed method takes reference images restricted to specific class groups as input and prunes
        (Steps~1 to 3 in Fig.~\ref{Fig:improved_method}, corresponding to lines~1 to 8 in Supplementary Information).
        In addition to the existing method, the proposed method includes two additional functions: "Change of pruning rate" and "Change of pruning order".
        This section explains the functions of "Change of pruning rate" and "Change of pruning order," respectively.
    
        \textbf{Change of pruning rate.}
            In the "Change of pruning rate" module, the post-pruning harmonic mean accuracy $A^{\prime}$ is compared with the pre-pruning value $A$. 
            If the $A'$ is maintained, the relevance is recalculated, and the bottom $P_{step}$\% of filters with the lowest relevance are pruned (l.9 to l.10 in Supplementary Information).
            Following the research by Yeom et al.\cite{YEOM}, the initial value for the increment of the pruning rate $P_{step}$  is set to 5\%. 
            If the $A'$ is not maintained, the process returns to Step 3, and the additional pruning rate $P_{step}$ is halved (l.10 to l.12 in Supplementary Information).
            If halving the pruning rate is repeated until there is only one filter to be pruned, the process transitions to "Change of pruning order" (l.13 in Supplementary Information).
    
        \textbf{Change of pruning order.}
            In "Change of pruning order," the filter with the lowest relevance is assigned a maximum relevance score to exclude it from the pruning targets, and the process returns to Step 3 (l.13 to l.20 in Supplementary Information).
            If accuracy still decreases even after excluding the filter with the lowest relevance $T_{max}$ times, the pruned model $M'$ with the highest $A'$ among the $T_{max}$ attempts is adopted (l.21 to l.23 in Supplementary Information).
            $T_{\mathrm{max}}$ can take values in the range $0 < T_{\mathrm{max}} < F_{\mathrm{num}}$ (where $F_{\mathrm{num}}$ is the total number of remaining filters).
            In this study, $T_{\mathrm{max}}$ was empirically determined to be 10.
            Preliminary results showed that this value, which is small relative to the total number of filters, is sufficient, as filters causing accuracy degradation despite low relevance are effectively excluded from the pruning candidates.
\section{Experiments and Discussion}\label{sec:experiment}
    This section presents comparative experiments demonstrating that DPX-SD, an LRP-based pruning method with dynamic control guided by class-accuracy metrics, effectively prunes filters of pre-trained CNNs while preserving domain-specific classification performance under data-scarce conditions.
    
    After describing the experimental settings, we present accuracy and compression performance on multiple ImageNet pre-trained CNNs, investigate the relationship between inter-class accuracy and pruning rate, and verify that the cascading accuracy degradation discussed in Sec.~\ref{sec:factor_countermeasure} is suppressed by the proposed method.
    Next, we analyze performance trends when varying the number of reference images and the number of target classes to assess the robustness of the method.
    We further validate design choices by examining the appropriateness of using the harmonic mean as an accuracy metric and by conducting a sensitivity analysis of $T_{\max}$.
    In addition, we evaluate generality by applying the same accuracy-guided control to a non-LRP pruning criterion (weight magnitude) and by pruning transfer-learned models on a medical imaging dataset.
    Finally, we isolate the contribution of each added mechanism via ablation studies and analyze computation time and additional iteration counts to discuss computational cost and practical acceptability.
    
    \subsection{Experimental Settings}\label{sec:experiment_setting}
    Following the experimental setup of existing research \cite{PXP,YEOM}, we evaluated the pruning of ImageNet-pretrained CNNs for specific subclasses by limiting the reference images, thereby simulating scenarios that necessitate transfer learning with scarce data.
    The target models are four ImageNet pre-trained CNNs: VGG16 (with and without BatchNorm) \cite{vgg}, ResNet-18, and ResNet-50 \cite{resnet}.
    For relevance calculation, images of 2 randomly selected classes from ImageNet (30 images per class) are used as reference images. 
    All reported results represent the average across five trials using different random seeds.
    The maximum pruning rate $P_{max}$ is set to 95\%.
    Under these conditions, we compare the proposed method with the LRP-based pruning method by Hatef et al. \cite{PXP} and the method by Yeom et al. \cite{YEOM}
    , and a weight-magnitude-based pruning method as a non-LRP baseline.
    In this paper, Hatef et al.'s method is referred to as Pruning by eXplain (PX), and Yeom et al.'s method is referred to as Dynamic Pruning by eXplain (DPX), and the weight-magnitude–based pruning method as Weight.
    
    \subsection{Overall Performance}\label{sec:overall_performance}
\begin{figure}[H]
            \centering
            \includegraphics[width=\columnwidth]{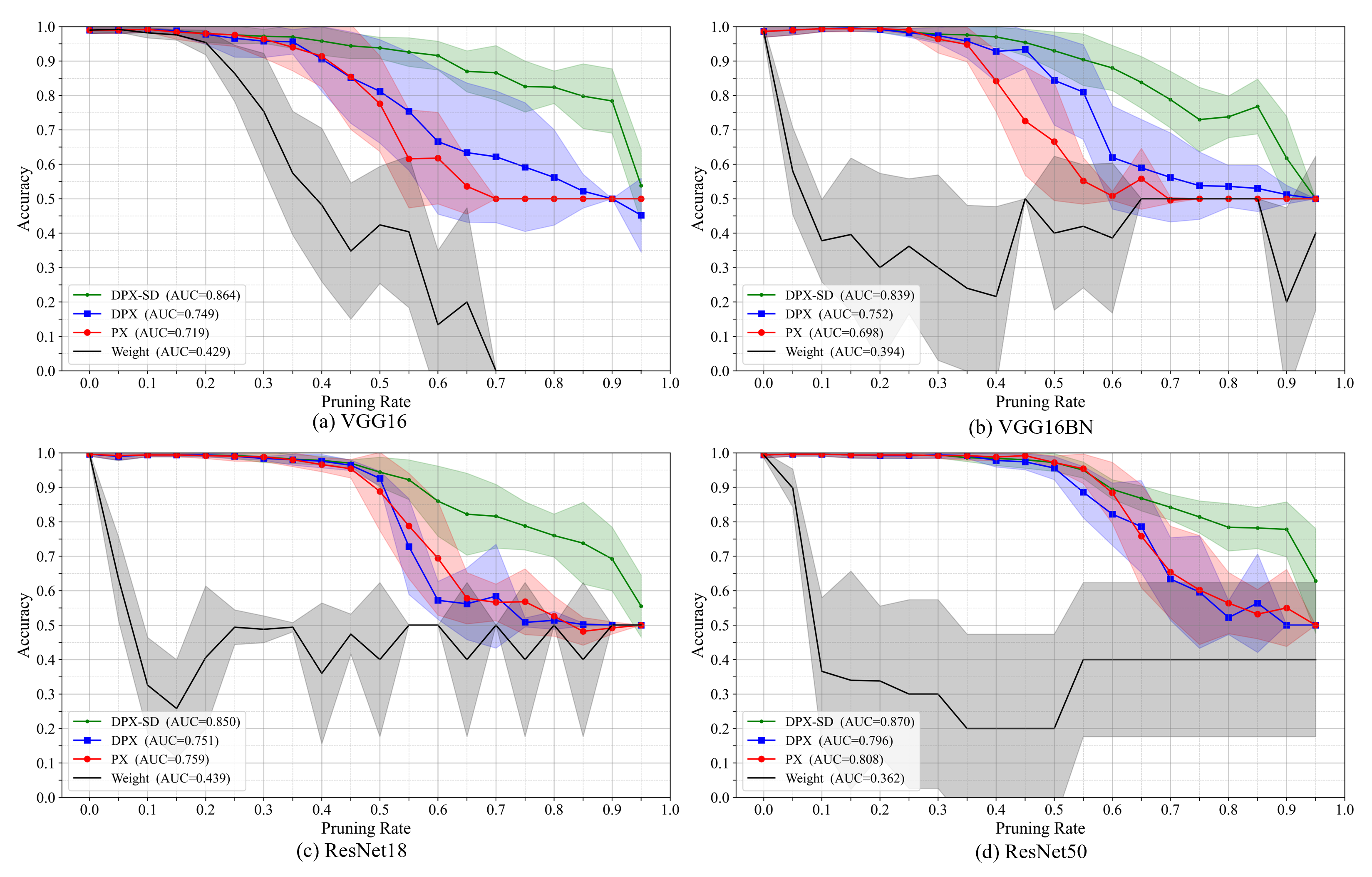}
            \caption{Relationship between accuracy and pruning rate for four CNN architectures: VGG16 (upper left), VGG16BN (upper right), ResNet-18 (lower left), and ResNet-50 (lower right). Green denotes the proposed method, while red, blue, and black denote existing methods.}
            \label{Fig:Acc_vs_pruning_rate}
        \end{figure}

        \textbf{Accuracy--Pruning Rate Curves and AUC.}
        Fig.~\ref{Fig:Acc_vs_pruning_rate} shows the relationship between accuracy and pruning rate for each CNN, where the shaded regions indicate the standard deviation across five trials.
        Table~\ref{AUC} summarizes the area under the accuracy--pruning rate curve (AUC) for each model.
        The AUC is computed using the trapezoidal rule on the accuracy-pruning rate curve up to $P_{\max}=0.95$, and normalized by $P_{\max}$:
        \begin{equation}
        \mathrm{AUC}
        =\frac{1}{P_{\max}}
        \sum_{i=1}^{N-1}
        \frac{\mathrm{Acc}_i + \mathrm{Acc}_{i+1}}{2}
        \,(p_{i+1}-p_i),
        \end{equation}
        where $p_i$ denotes the $i$-th pruning rate, $\mathrm{Acc}_i$ is the corresponding classification accuracy, $N$ is the number of sampled pruning points, and $P_{\max}$ is the maximum pruning rate used for evaluation.
        As shown in Table~\ref{AUC}, the proposed pruning control mechanism tends to maintain higher accuracy throughout pruning than the baselines across all models.
        DPX~\cite{YEOM} also re-explores the pruning order by recalculating relevance, similarly to the proposed method.
        However, compared to PX~\cite{PXP}, DPX sometimes improves performance but can also degrade it depending on the setting.
        This behavior is likely attributable to the cascading accuracy degradation triggered by imbalances in class-wise accuracy (see Sec.~\ref{sec:factor_countermeasure}).
        In Fig.~\ref{Fig:Acc_vs_pruning_rate}, the accuracy gap between the proposed and existing methods widens as the pruning rate increases, and it narrows only as the pruning rate approaches 100\%.
        This suggests that the accuracy-preserving effects of ``Change of pruning rate'' and ``Change of pruning order'' are substantial during iterative pruning, whereas the fundamental loss of representational capacity dominates at extremely high pruning rates.

        \begin{table}[H]
            \centering
            \caption{Class-averaged area under the accuracy--pruning rate curve (AUC) for each ImageNet pre-trained CNN. Higher is better.}
            \label{AUC}
        \begin{tabular}{lcccc}
                \toprule
                Model & DPX-SD & DPX & PX & Weight \\
                \midrule
                VGG16     & \textbf{0.864} & 0.749 & 0.719 & 0.429 \\
                VGG16BN   & \textbf{0.839} & 0.752 & 0.698 & 0.394 \\
                ResNet-18 & \textbf{0.850} & 0.751 & 0.759 & 0.439 \\
                ResNet-50 & \textbf{0.870} & 0.796 & 0.808 & 0.362 \\
                \bottomrule
            \end{tabular}
        \end{table}
    
\begin{figure}[H]
            \centering
            \includegraphics[width=0.75\columnwidth]{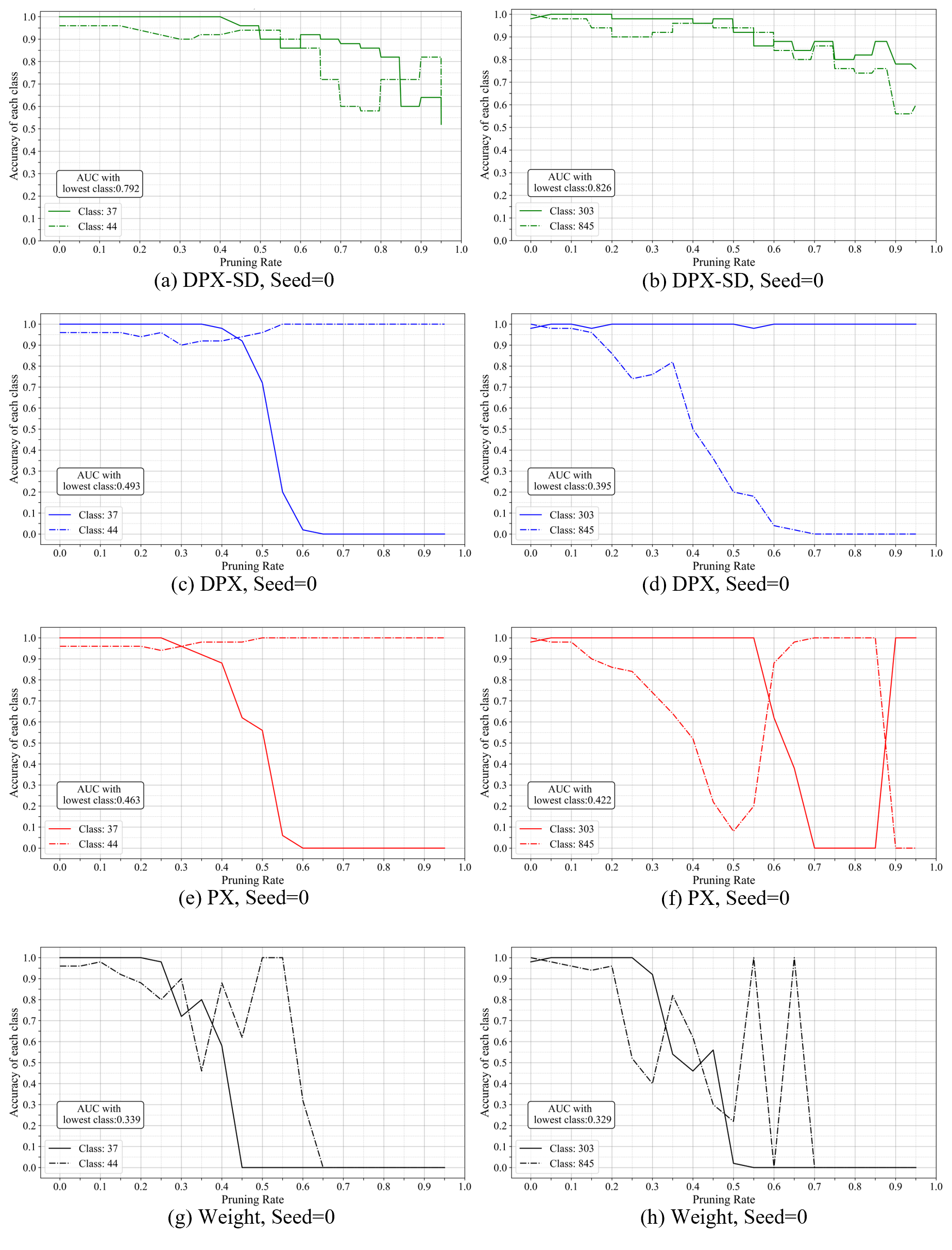}
            \caption[Relationship between accuracy and pruning rate for each class in VGG16 pruned by Weight (Left: Seed=0, Right: Seed=1)]
            {Relationship between accuracy and pruning rate for each class in VGG16 pruned by each method (Left: Seed=0, Right: Seed=1)}
            \label{Fig:Acc_vs_pruning_rate_per_class}
        \end{figure}

        \textbf{Each Class Accuracy and Lowest-Class AUC.}
        To investigate whether the proposed pruning control mechanism can suppress class-specific accuracy collapse and maintain balanced performance across classes during pruning, we analyze the class-wise accuracy transitions.
        Fig.~\ref{Fig:Acc_vs_pruning_rate_per_class} shows the class-wise accuracy transitions in VGG16 during pruning.
        The left and right panels correspond to different random seeds, and each plot reports accuracy for each class.
        In each graph, ``AUC with lowest class'' denotes the area under the curve of the minimum class accuracy at each pruning rate; a larger value indicates that extreme accuracy drops in specific classes are avoided.
        Table~\ref{AUC_lowest_class} summarizes the ``AUC with lowest class'' for each method and seed.
        As shown in Table~\ref{AUC_lowest_class}, the proposed pruning control mechanism suppresses accuracy degradation in specific classes more effectively than existing methods.
        From the plots, with existing methods (PX, DPX), once a specific class starts to degrade, its accuracy rapidly collapses, ultimately resulting in a model that predicts only a single class.
        Furthermore, with the weight-magnitude-based pruning method (Weight), the accuracy of all classes eventually converges to nearly zero, indicating that signals are no longer properly propagated to the output layer.
        In contrast, the method with the pruning control mechanism added (DPX-SD) visually exhibits suppression of class-specific accuracy collapse.
        This trend is also supported by the fact that the proposed pruning control mechanism achieves the highest ``AUC with lowest class'' across seeds.
        
        \begin{table}[H]
            \centering
            \caption{AUC with lowest class on VGG16, computed as the area under the minimum class-accuracy curve across pruning rates. Higher values indicate better suppression of class-specific accuracy collapse.}
            \label{AUC_lowest_class}
            \begin{tabular}{lcccc}
                \toprule
                Seed & DPX-SD & DPX & PX & Weight \\
                \midrule
                Seed 0 & \textbf{0.792} & 0.493 & 0.463 & 0.339 \\
                Seed 1 & \textbf{0.826} & 0.395 & 0.422 & 0.329 \\
                \bottomrule
            \end{tabular}
        \end{table}
    
    \subsection{Robustness under Data-Scarce Conditions}\label{sec:robustness}
\begin{figure}[H]
            \centering
            \includegraphics[width=0.65\columnwidth]{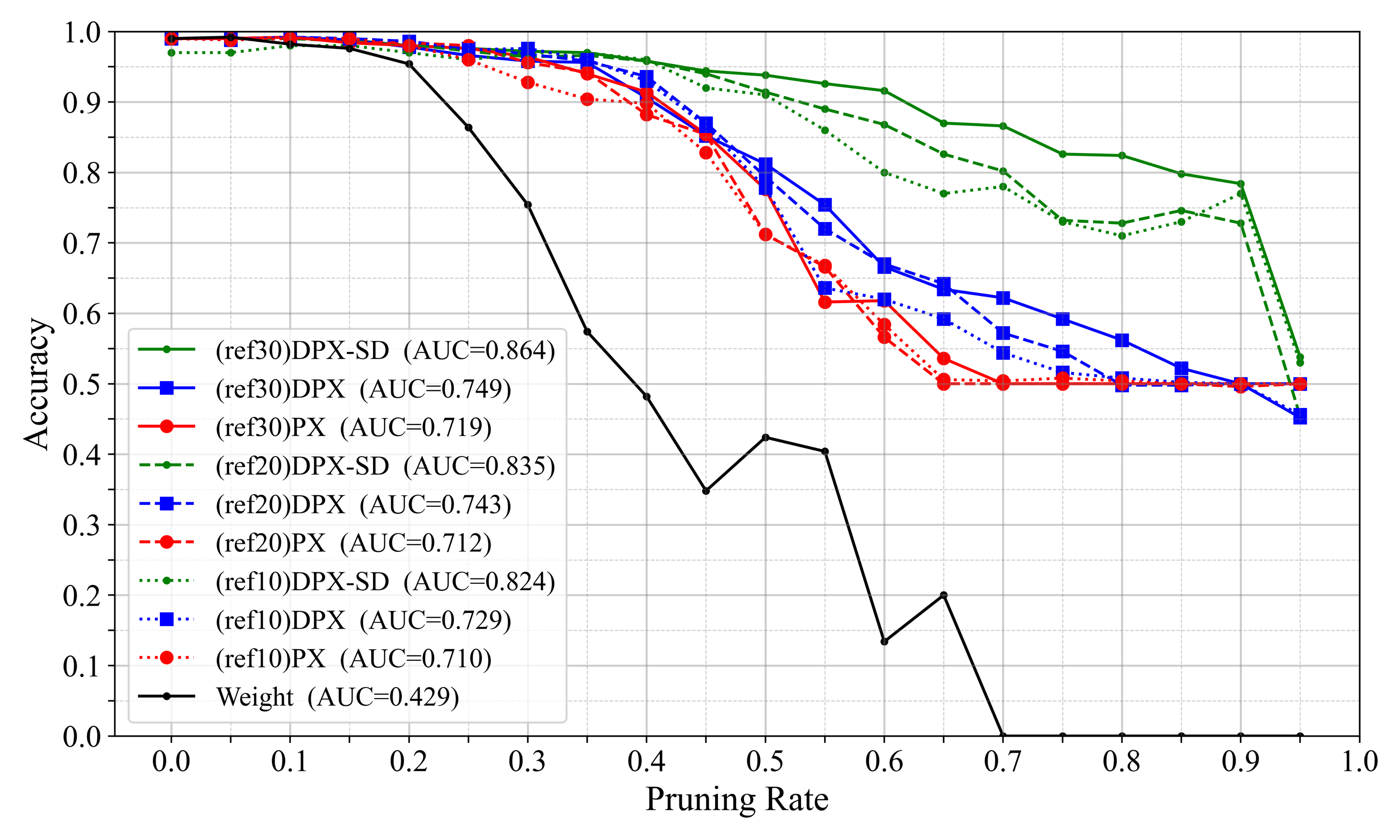}
            \caption[Relationship between accuracy and pruning rate when varying the number of reference images in VGG16]
            {Relationship between accuracy and pruning rate when varying the number of reference images in VGG16}
            \label{Fig:Acc_vs_pruning_rate_each_ref_num}
        \end{figure}

        \textbf{Varying the Number of Reference Images.}
        We examined how the accuracy--pruning rate relationship changes when varying the number of reference images in VGG16.
        As shown in Fig.~\ref{Fig:Acc_vs_pruning_rate_each_ref_num}, the method with the pruning control mechanism added (DPX-SD) maintains higher accuracy during pruning as the number of reference images increases.
        In the 60--90\% pruning range, even with only 10 reference images, the method with the pruning control mechanism added maintains higher accuracy than the baseline using 30 reference images.
        These results indicate that the proposed pruning control mechanism can effectively retain critical classification filters even with fewer reference images.

\begin{figure}[H]
            \centering
            \includegraphics[width=0.65\columnwidth]{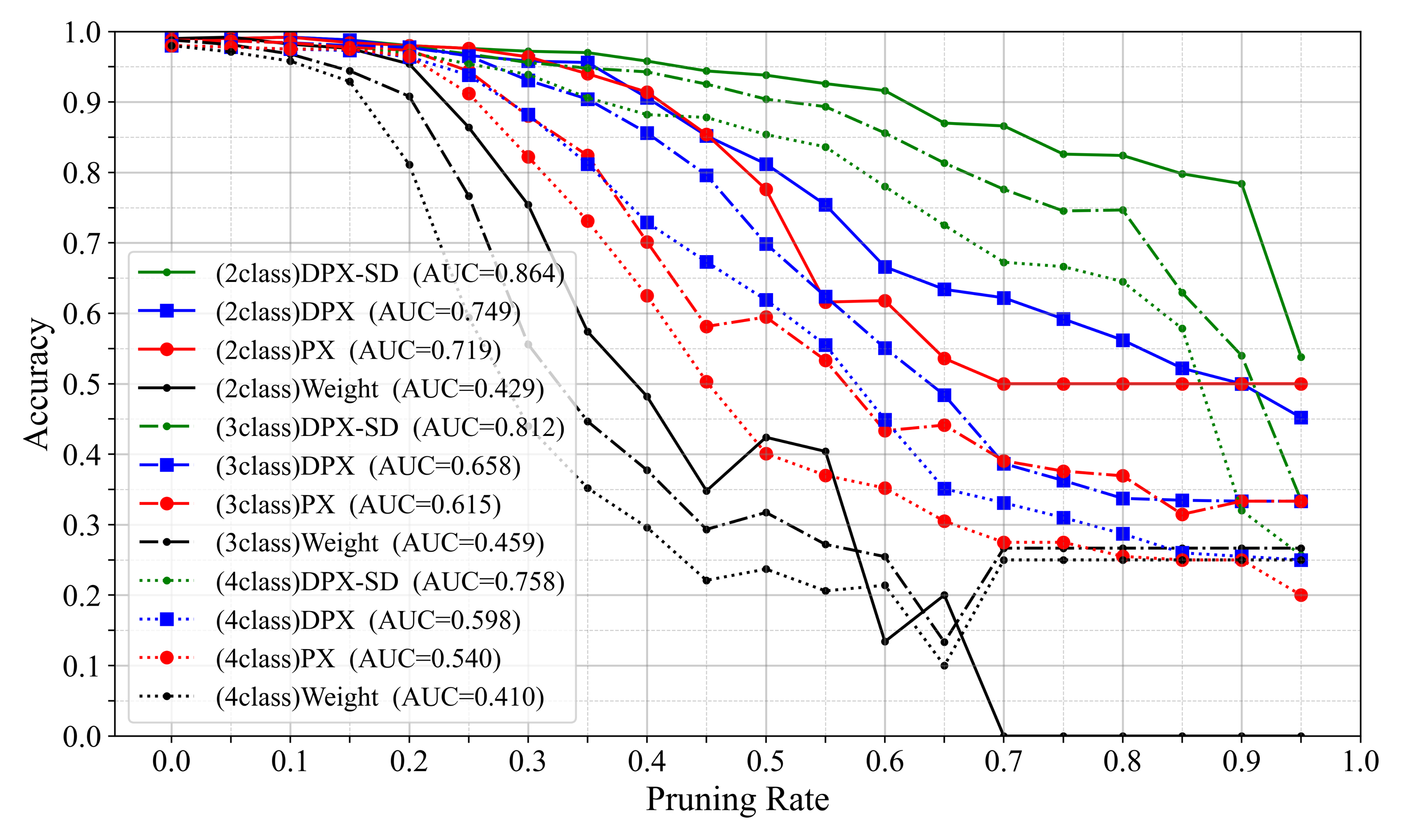}
            \caption[Relationship between accuracy and pruning rate when varying the number of classes in VGG16]
            {Relationship between accuracy and pruning rate when varying the number of limited classes in VGG16}
            \label{Fig:Acc_vs_pruning_rate_each_class_num}
        \end{figure}

        \textbf{Varying the Number of Classes.}
        Fig.~\ref{Fig:Acc_vs_pruning_rate_each_class_num} shows the relationship between accuracy and pruning rate when varying the number of classes.
        At the same pruning rate, increasing the number of classes degrades accuracy across all methods.
        This is because a larger class set increases representational demands, limiting the number of filters that can be pruned.
        Nevertheless, the method with the pruning control mechanism added (DPX-SD) consistently preserves higher accuracy than baselines for all class counts.

    \subsection{Design Validation}\label{sec:design_validation}
\begin{figure}[H]
            \centering
            \includegraphics[width=0.65\columnwidth]{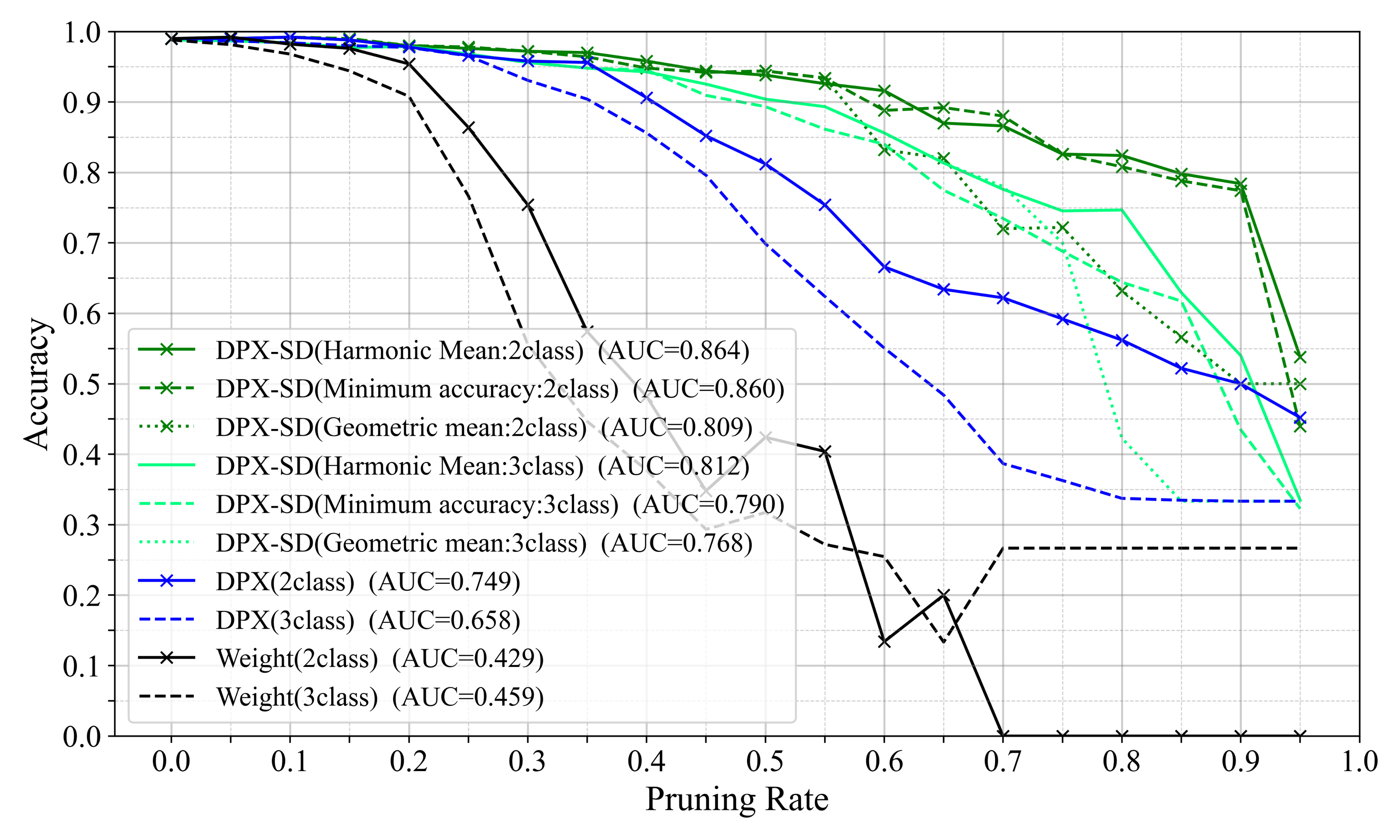}
            \caption{
            Accuracy-pruning-rate curves in VGG16 when the pruning control metric is changed from the harmonic mean to the geometric mean and the minimum class accuracy.}
            \label{Fig:GEO_MIN}
        \end{figure}

        \textbf{Choice of Accuracy Metric.}
        The proposed pruning control mechanism adjusts the pruning rate and pruning order based on a single scalar metric that aggregates class-wise accuracies. 
        We adopt the harmonic mean of class accuracies as this metric and evaluate its suitability by comparing it with the geometric mean and the minimum class accuracy. 
        Fig.~\ref{Fig:GEO_MIN} shows the results when the control metric is replaced with these alternatives.
        In terms of AUC, the harmonic mean and the minimum class accuracy exhibit nearly identical performance in the two-class setting, whereas the geometric mean results in a lower AUC due to its weaker sensitivity to accuracy drops. 
        When extended to three classes, however, the harmonic mean outperforms the minimum class accuracy, demonstrating more stable behavior in multi-class conditions. 
        While the minimum class accuracy focuses only on the worst-performing class, the harmonic mean reflects all classes while remaining sensitive to low accuracies, thereby capturing both overall performance and class balance. 
        For these reasons, we adopt the harmonic mean as the control metric, as it provides the most balanced trade-off between worst-class protection, overall performance, and scalability to multi-class settings.

\begin{figure}[H]
            \centering
            \includegraphics[width=0.65\columnwidth]{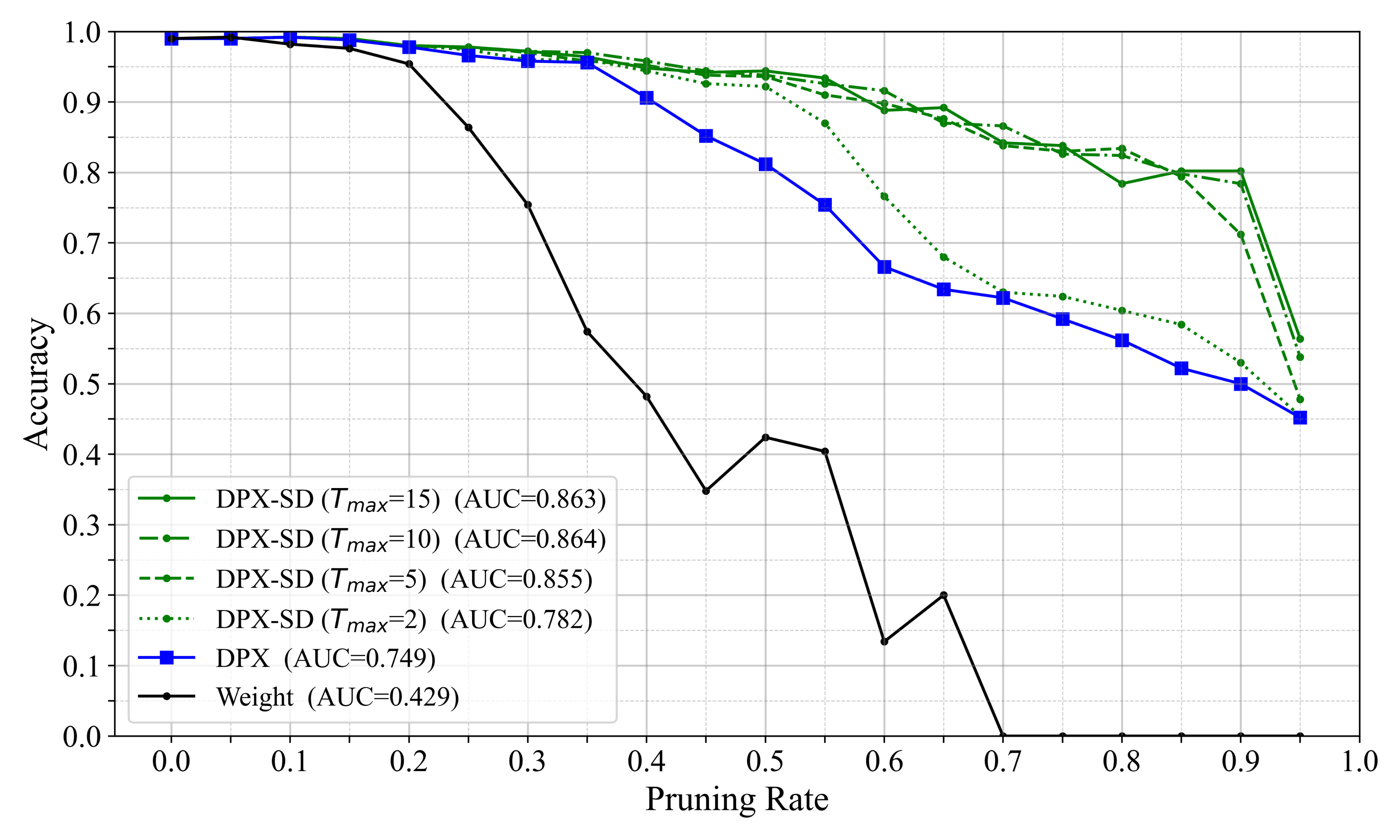}
            \caption
            {Accuracy-pruning-rate curves in VGG16 when varying the maximum number of trials 
            $T_{max}$ in the change-of-pruning-order mechanism.}
            \label{Fig:AUC_Tmax}
        \end{figure}

        \textbf{Sensitivity Analysis for Trial Count.}
        In the ``Change of pruning order'' mechanism, we set an upper bound $T_{\max}$ on the number of trials, 
        which directly affects both the search scope and the computational cost; therefore, we conduct a sensitivity analysis.
        As shown in Fig.~\ref{Fig:AUC_Tmax}, $T_{\max}=2$ results in a clear degradation of AUC, 
        indicating that an insufficient number of trials fails to avoid filters that cause accuracy collapse. 
        Increasing $T_{\max}$ from $5$ to $10$ improves the AUC, 
        whereas the results for $10$ and $15$ are nearly identical.
        This tendency indicates that the performance saturates around $T_{\max}=10$, 
        and that most harmful filters are already avoided within this range. 
        We adopt $T_{\max}=10$ in this study, as it provides near-maximum performance without incurring unnecessary additional computation.

    \subsection{Generality}\label{sec:generality}
\begin{figure}[H]
            \centering
            \includegraphics[width=0.65\columnwidth]{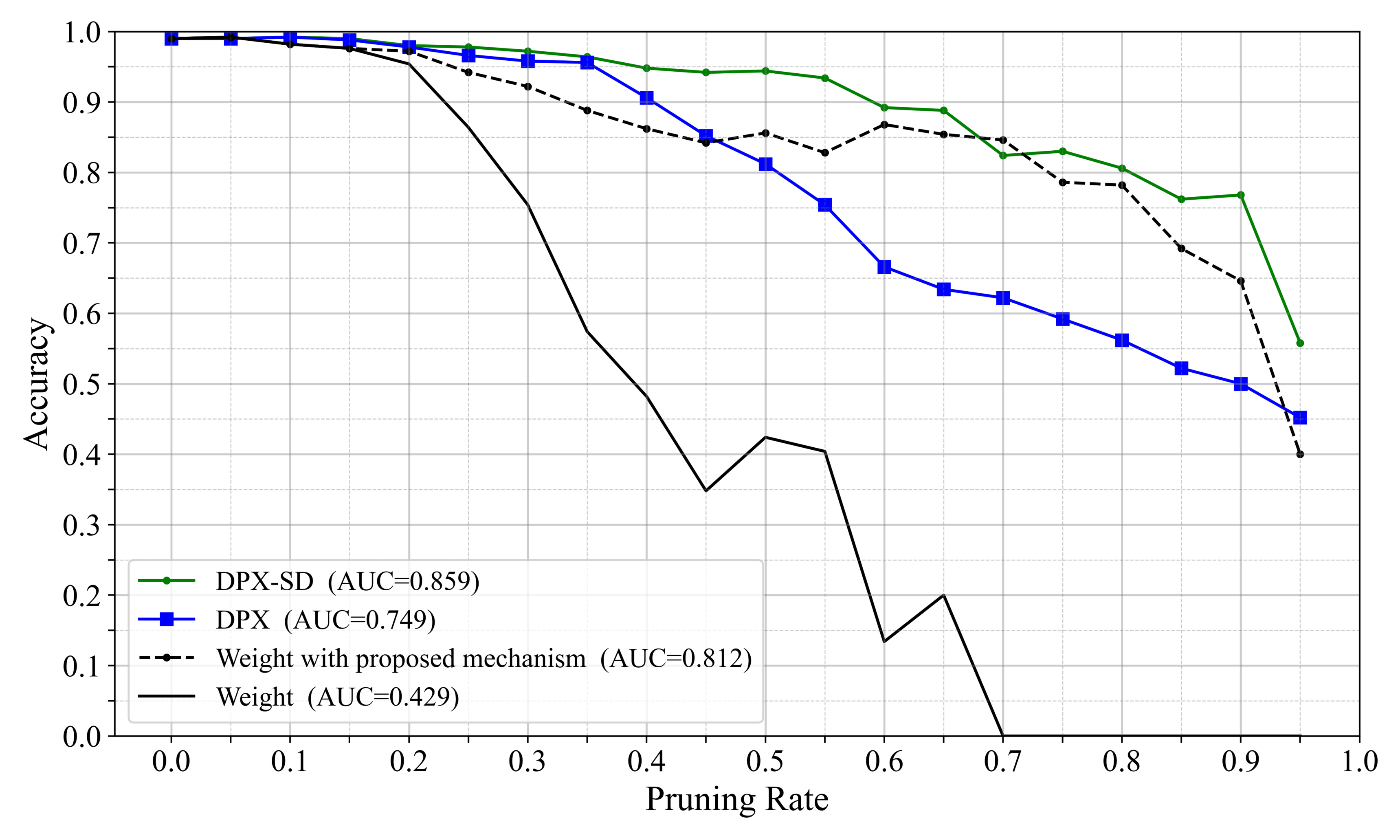}
            \caption{Effect of applying the accuracy-guided control mechanism to relevance- and weight-based pruning in VGG16.}
            \label{Fig:DPW-SD}
        \end{figure}

        \textbf{Applying the Control Mechanism to a Non-LRP Criterion (Weight Magnitude).}
        To verify the effectiveness of the accuracy-guided control mechanism itself, we apply the same control to a representative non-LRP pruning criterion, weight magnitude.
        Fig.~\ref{Fig:DPW-SD} compares the baseline pruning by weight magnitude and relevance, and their controlled variants augmented with the accuracy-guided mechanism.
        Even when applied to weight magnitude, the same control improves AUC from 0.429 to 0.812, indicating that the idea is not specific to LRP.
        However, the controlled relevance-based pruning achieves a higher AUC (0.864) than the controlled weight-based pruning (0.812).
        This is likely because weight magnitude reflects only a static parameter norm, whereas relevance captures contribution to model outputs during inference, leading to more discriminative pruning decisions.

        \textbf{Real-World Transfer Learning (Medical Imaging).}
        To confirm that the effectiveness of the proposed method is not limited to ImageNet subclasses,
        we conduct an additional experiment on a small real-world dataset, the Pneumonia X-ray dataset\cite{kermany2018identifying}.
        In this experiment, we formulate a binary classification task (pneumonia vs. normal).
        An ImageNet pre-trained CNN (VGG16) is used as a feature extractor,
        and only the final fully connected layer is trained as a classifier on the target dataset.
        We then apply Weight, PX, DPX, and DPX-SD to prune the CNN component of this transfer-learned model.
        For relevance and pruning control, we use 30 reference images per class,
        while all other experimental settings are kept identical to those in the previous ImageNet subclass experiments.
        As shown in Fig.~10, the same ordering in AUC is observed on this dataset,
        namely DPX-SD > DPX > PX > Weight.
        This result suggests that the proposed accuracy-feedback pruning control is not limited to ImageNet subclasses.

        \begin{figure}[H]
            \centering
            \includegraphics[width=0.65\columnwidth]{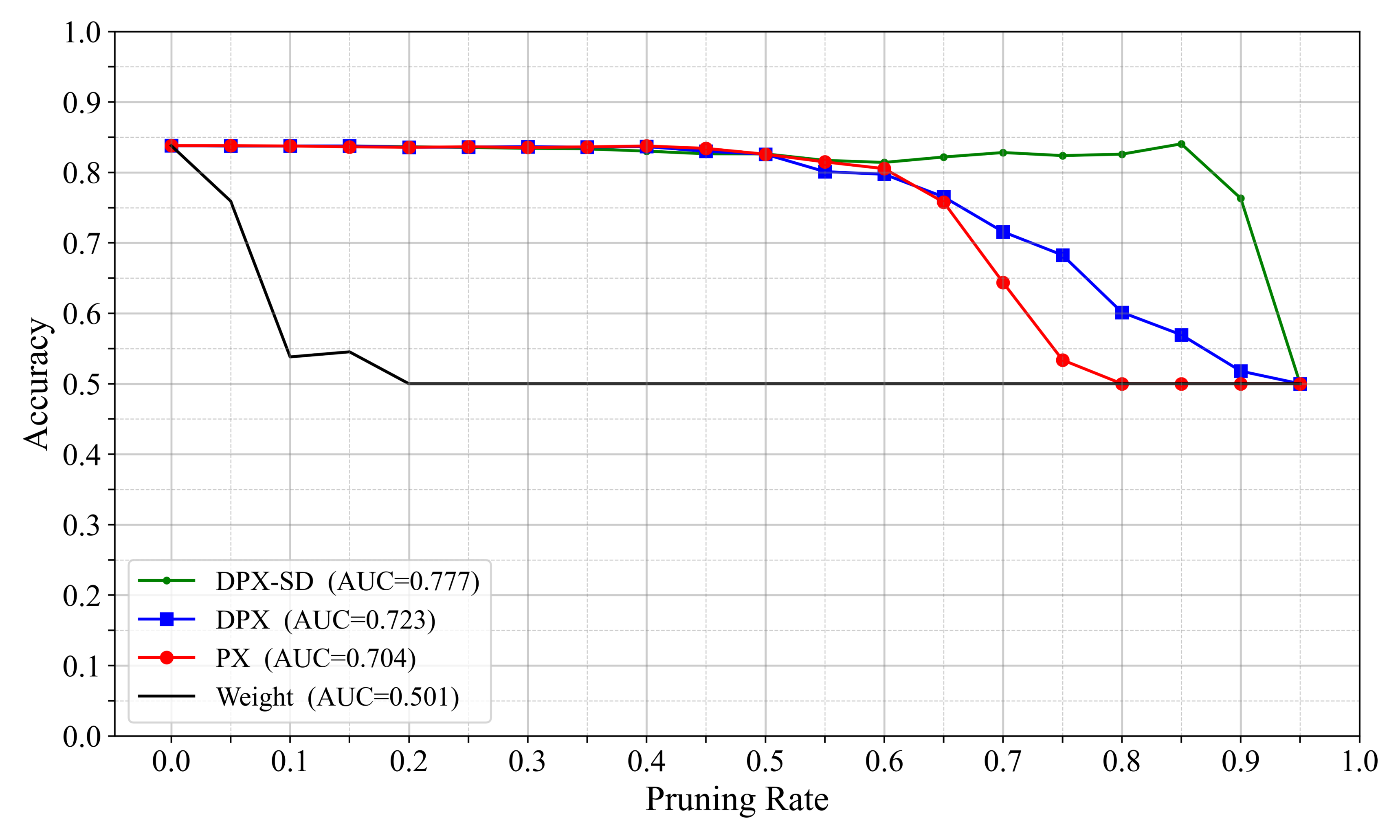}
            \caption
            {Accuracy-pruning rate curves on the Pneumonia X-ray dataset using a transfer-learned VGG16. DPX-SD preserves higher accuracy than existing pruning methods, indicating robustness of the accuracy-guided control mechanism in real-world transfer learning.}
            \label{Fig:PNEUMONIA}
        \end{figure}
    
    \subsection{Ablation Study}\label{sec:ablation}
\begin{figure}[H]
        \centering
        \includegraphics[width=0.65\columnwidth]{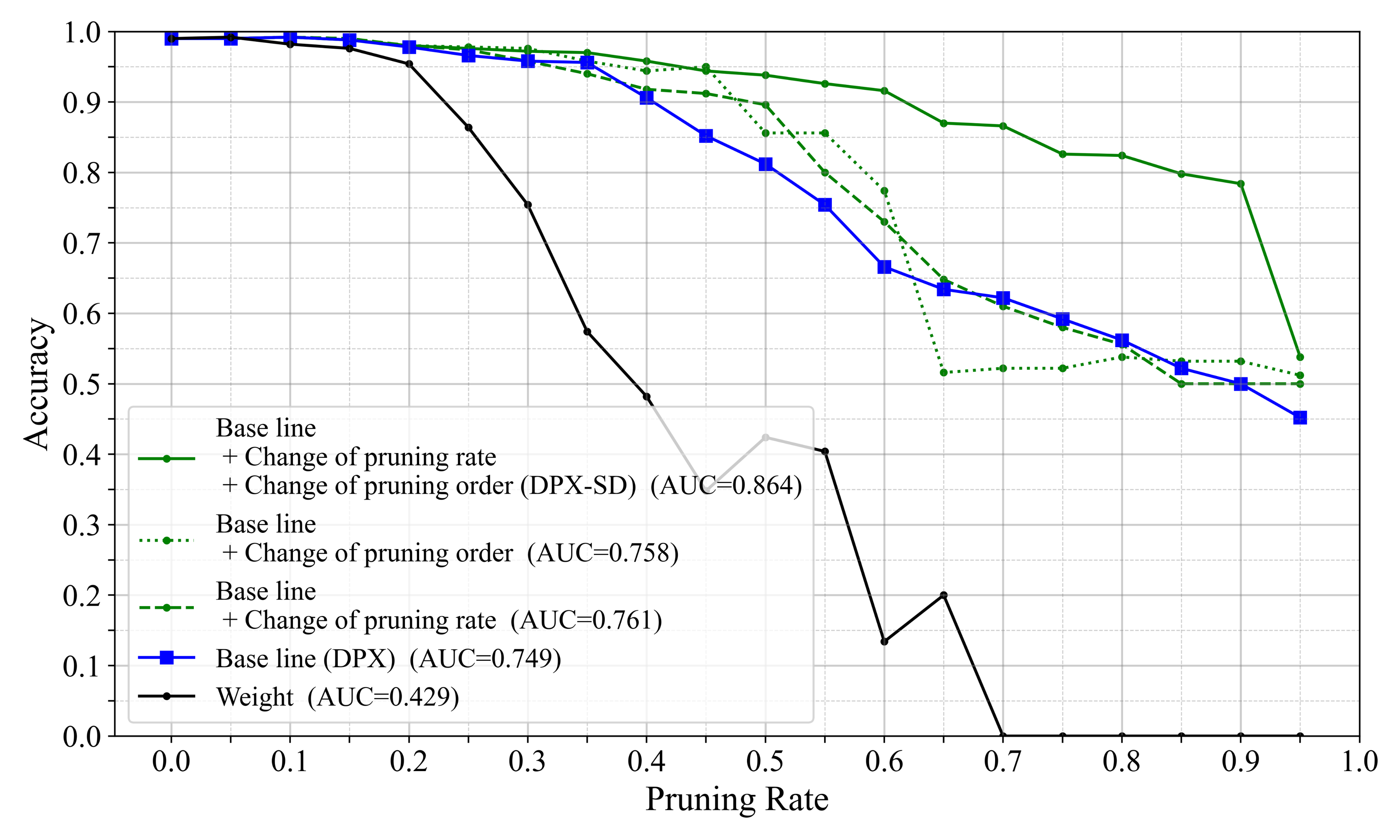}
        \caption
        {Accuracy-pruning-rate curves for the ablation study on VGG16 using ImageNet subclasses. Each variant isolates the effect of the proposed control mechanisms (“Change of pruning rate” and “Change of pruning order”).}
        \label{Fig:abblation_study}
    \end{figure}

    To clarify the individual contribution of each component described in Sec.~\ref{sec:Proposed_method}, we conduct an ablation study (Fig.~\ref{Fig:abblation_study}).
    Adding only ``Change of pruning order'' or only ``Change of pruning rate'' to the baseline improves AUC compared to the baseline.
    However, the improvement from either component alone is limited compared to using both together (DPX-SD).
    This is because ``Change of pruning rate'' suppresses abrupt accuracy drops due to overly aggressive batch pruning, while ``Change of pruning order'' prevents accidental removal of critical filters among low-relevance candidates; the two play complementary roles.

    \subsection{Computational Cost}\label{sec:computational_cost}
\begin{figure}[H]
        \centering
        \includegraphics[width=\columnwidth]{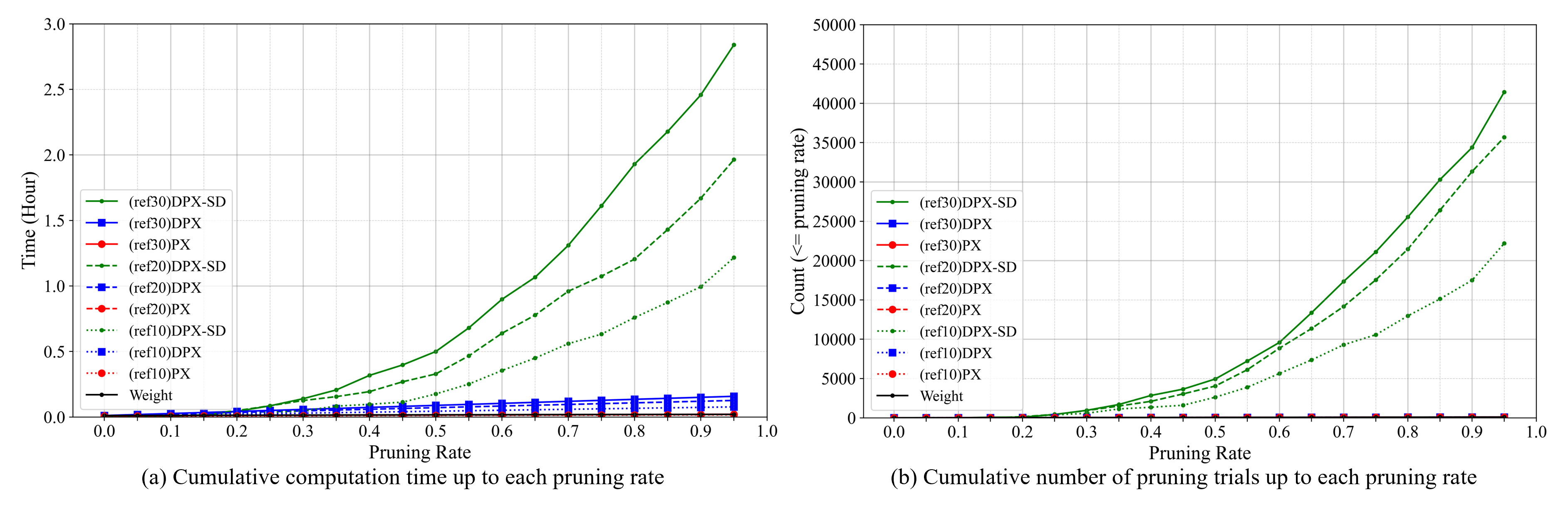}
        \caption[Relationship between computation time and pruning rate when varying the number of reference images in VGG16]
        {Left: Computation time vs. pruning rate under varying numbers of reference images (VGG16). Right: Number of pruning iterations vs. pruning rate under the same settings.}
        \label{Fig:time_count}
    \end{figure}

    Fig.~\ref{Fig:time_count} illustrates the relationship between the pruning rate and both the cumulative computation time and the cumulative number of pruning trials under different numbers of reference images. 
    As the number of reference images increases, both the cumulative computation time and the cumulative number of pruning trials grow. 
    In particular, the left and right graphs exhibit nearly similar shapes, indicating that the increase in computation time closely corresponds to the increase in the number of pruning iterations.

    This tendency can be interpreted as follows. 
    Increasing the number of reference images makes accuracy degradation easier to detect, which more frequently triggers reductions of the additional pruning rate and re-evaluations of the pruning order. 
    In other words, a larger reference set leads to more extensive exploration during pruning, and the resulting increase in iteration count directly contributes to the growth of total computation time.
    
    However, this additional computational cost is incurred mainly during a one-time offline compression procedure and does not affect inference after pruning. 
    Therefore, it is acceptable in applications where high inference accuracy and stability are prioritized and model updates are infrequent, such as medical image diagnosis or satellite image analysis. 
    In contrast, in environments that require frequent retraining and recompression due to distribution shifts, or in time-critical workflows such as online learning and in-training model updates in reinforcement learning, the additional computational overhead may become a practical limitation.
\section{Conclusion}
    We proposed DPX-SD, an accuracy-aware pruning control framework that suppresses cascading accuracy degradation in LRP-based CNN pruning under data-scarce transfer learning. 
    By dynamically adjusting the pruning rate and re-evaluating the pruning order based on the harmonic mean of class accuracies, DPX-SD consistently outperformed existing LRP-based methods across four CNN architectures, improving the class-averaged AUC of VGG16 by approximately 15\%. 
    The framework is also effective when applied to non-LRP criteria and real-world medical imaging tasks. 
    
    However, since DPX-SD iteratively refines the pruning order to maintain accuracy, it incurs higher computational costs than existing methods. 
    Future work will focus on improving computational efficiency by optimizing relevance updates, selecting pruning candidates more effectively, and introducing approximate calculations. 

\section*{Data Availability}
    The datasets and code used during the current study are available from the corresponding author on reasonable request.

\section*{Funding}
    This work was supported by JSPS KAKENHI Grant Numbers JP22K17969.
\bibliographystyle{plain}
\bibliography{sample}

\section*{Supplementary Information: Pseudo-code of DPX-SD}
    Algorithm \ref{algo:DPX-SD} presents the pseudo-code for the proposed method described in Sec. 3.2.
    The auxiliary functions appearing within the pseudo-code are described in the "Details of Auxiliary Functions" section.

    \begin{algorithm}
    \caption{DPX-SD: Dynamic Pruning by eXplain for Scarce Data}
    \begin{algorithmic}
    \State \textbf{Input:} CNN model $\mathbf{M}$, filter total number $F_{num}$, reference images $\mathbf{X}_r$, reference labels $\mathbf{Y}_r$, test images $\mathbf{X}_t$, test labels $\mathbf{Y}_t$, pruning step $P_{step}\%$, max total skips $T_{max}$, max pruning rate $P_{max}\%$
    \State 1: \ $P = 0$, $P_{step}' = P_{step}$ \Comment{Initial pruning rate and step}
    \State 2: \ $A = \text{EvaluateFunction}(\mathbf{M}, \mathbf{X}_r, \mathbf{Y}_r)$\Comment {Evaluate $A$}
    \State 3: \ \textbf{while} $P < P_{max}$ \textbf{:}
    \State 4: \ \hspace{1em} $\mathbf{S} = [0, 0, \dots, 0]$\Comment{Initialize a vector of the sum of relevance for each filter (length $F_{\mathrm{num}}$)}
    \State 5: \ \hspace{1em} $P' = P + P_{step}'$\Comment{Temporary update of pruning rate}
    \State 6: \ \hspace{1em} \textbf{for} $\mathbf{x}_r$ in $\mathbf{X}_r$ \textbf{:}
    \State 7: \ \hspace{2em} $\mathbf{R}_{\mathbf{x}_r} = \text{LRP}(\mathbf{M}, \mathbf{x}_r)$, $\mathbf{S} = \mathbf{S} + \mathbf{R}_{\mathbf{x}_r}$ \Comment{Step 1 \& 2: Get and sum relevance}
    \State 8: \ \hspace{1em} $\mathbf{M'} = \text{FilterPruner}(\mathbf{M}, \mathbf{S}, P')$\Comment{Step 3 : Remove $P'$\% lowest relevance filter}
    \State 9: \ \hspace{1em} $A' = \text{EvaluateFunction}(\mathbf{M'}, \mathbf{X}_r, \mathbf{Y}_r)$
    \State 10:\ \hspace{1em} \textbf{if} $0 < (A - A')$ \textbf{:}\Comment{Start \textbf{"Change of pruning rate"}(See Sec. 3.2.1)}
    \State 11:\ \hspace{2em} \textbf{if} $1 < \text{INT}(P_{step}'F_{num})$ \textbf{:}\Comment{\text{INT}($P_{step}'F_{num}$) is number of filters to be pruned}
    \State 12:\ \hspace{3em} $P_{step}' = P_{step}'/2$\Comment{Reduce additional pruning rate by half}
    \State 13:\ \hspace{2em} \textbf{else:}\Comment{Start \textbf{"Change of pruning order"}(See Sec. 3.2.2)}
    \State 14:\ \hspace{3em} $T = 0$, $\mathbf{D} = \{\}$\Comment{Initialize total skip count and save dictionary}
    \State 15:\ \hspace{3em} \textbf{while} $0 < (A - A')$ \textbf{:}
    \State 16:\ \hspace{4em} \textbf{if} $T < T_{max}$ \textbf{:}
    \State 17:\ \hspace{5em} $\mathbf{S'} = \text{BoostLowRelevance}(\mathbf{S}, T)$\Comment{Maximize lower $T$ relevances}
    \State 18:\ \hspace{5em} $\mathbf{M'} = \text{FilterPruner}(\mathbf{M}, \mathbf{S'}, P')$, $A' = \text{EvaluateFunction}(\mathbf{M'}, \mathbf{X}_r, \mathbf{Y}_r)$
    \State 19:\ \hspace{5em} $\mathbf{D}[A'] = M'$\Comment{Register $M'$ in $D$ with $A'$ as the key}
    \State 20:\ \hspace{5em} $T = T + 1$\Comment{Count up $T$}
    \State 21:\ \hspace{4em} \textbf{else:}
    \State 22:\ \hspace{5em} $\mathbf{M'}, A' = \text{BestModelChoice}(D)$\Comment{Choose best $A'$ model from $D$}
    \State 23:\ \hspace{5em} \textbf{break}
    \State 24:\ \hspace{1em} $\mathbf{M} = \mathbf{M'}$, $A = A'$, $P = P'$, $P'_{step} = P_{step}$\Comment{Update of various variables}
    \State 25:\ \hspace{1em} $Acc = \text{TestFunction}(\mathbf{M}, \mathbf{X}_t, \mathbf{Y}_t)$\Comment{Test pruned model and check accuracy}
    \State 26:\ \hspace{1em} print($Acc$)
    \State 27:\ \textbf{return} $\mathbf{M}$
    \end{algorithmic}\label{algo:DPX-SD}
    \end{algorithm}
    
    \subsection*{Details of Auxiliary Functions}\label{sec:AUX}
    
    \paragraph{FilterPruner($\bf{M}, \bf{S}, P'$)}
    This function takes the model $\mathbf{M}$ and the sum of relevance vectors $\mathbf{S}$ for all filters as input, and generates a new model $\mathbf{M'}$ by removing the bottom $P'\%$ of filters with the lowest relevance scores.
    
    \paragraph{BoostLowRelevance($\mathbf{S}, T$)}
    This function significantly increases the relevance scores of the $T$ filters with the lowest values in the sum of relevance vectors $\mathbf{S}$, thereby temporarily excluding them from pruning.
    This process is utilized in the "Change of pruning order" described in Sec.3.2.2.
    
    \paragraph{BestModelChoice($\mathbf{D}$)}
    This function receives a dictionary structure $\mathbf{D}$, which stores the evaluation value $A'$ as keys and the pruned models as values, as input, and selects the model $\mathbf{M'}$ that exhibits the highest evaluation value from among them.

    \paragraph{EvaluateFunction($\mathbf{M}$, $\mathbf{X}_r$, $\mathbf{Y}_r)$}
    This function computes the harmonic mean accuracy $A$ across all classes using the reference dataset ($\mathbf{X}_r, \mathbf{Y}_r$), as defined in Eq. 3. 
    
    \paragraph{TestFunction($\mathbf{M}$, $\mathbf{X}_t$, $\mathbf{Y}_t)$}
    This function evaluates the generalization performance of the pruned model $\mathbf{M}$ using the independent test dataset ($\mathbf{X}_t, \mathbf{Y}_t$). 
    It calculates and returns the standard top-1 classification accuracy. 
    This value is used to report the final performance of the model at each pruning step (e.g., for plotting accuracy curves).

\end{document}